# Design of calibration experiments for identification of manipulator elastostatic parameters

Alexandr Klimchik[1, 2] , Anatol Pashkevich[1, 2], Yier Wu[1, 2], Stéphane Caro[2, 3], Benoît Furet[2, 3]

*1 Ecole des mines de Nantes, Nantes (EMN), 44300, France*
*2 Institut de Recherche en Communications et Cybernétique de Nantes (IRCCyN), Nantes, 44300, France,*
*3 Centre National de la Recherche Scientifique (CNRS), France*
*4 Université de Nantes, Nantes, 44000, France*

**Abstract:** The paper is devoted to the elastostatic calibration of industrial robots, which are used for precise machining of large-dimensional parts made of composite materials. In this technological process, the interaction between the robot and the workpiece causes essential elastic deflections of the manipulator components that should be compensated by the robot controller using relevant elastostatic model of this mechanism. To estimate parameters of this model, an advanced calibration technique is applied that is based on the non-linear experiment design theory, which is adopted for this particular application. In contrast to previous works, it is proposed a concept of the user-defined test-pose, which is used to evaluate the calibration experiments quality. In the frame of this concept, the related optimization problem is defined and numerical routines are developed, which allow to generate optimal set of manipulator configurations and corresponding forces/torques for a given number of the calibration experiments. Some specific kinematic constraints are also taken into account, which insure feasibility of calibration experiments for the obtained configurations and allow avoiding collision between the robotic manipulator and the measurement equipment. The efficiency of the developed technique is illustrated by an application example that deals with elastostatic calibration of the serial manipulator used for robot-based machining.

## 1. Introduction

In the usual engineering practice, the accuracy of an anthropomorphic manipulator depends on many factors. In accordance with [1-2], the main sources of robot positioning errors can be divided into two main groups: geometrical (link lengths, assembling errors, errors in the joint zero values et al.) and non-geometrical ones (compliant errors, measurement errors, environment factors, control errors, friction, backlash, wear et al.). For the industrial manipulators, the most essential of them are related to the manufacturing tolerances leading to the geometrical parameters deviation with respect to their nominal values (the geometrical errors) as well as to the end-effector deflections caused by the applied forces and torques (the compliance errors). It is worth mentioning that these error sources may be either independent or correlated, but, in practice, they are usually treated sequentially, assuming that they are statistically independent.

Usually, for the industrial applications where the external forces/torques applied to the end-effector are relatively small, the prime source of the manipulator inaccuracy is the geometrical errors. As reported by several authors [3], they are responsible for about 90% of the total position error. These errors are associated with the differences between the nominal and actual values of the link/joint parameters. Typical examples of them are the differences between the nominal and the actual length of links, the differences between zero values of actuator coordinates in the real robot and the mathematical model embedded in the controller (joint offsets) [4]. They can be also induced by the non-perfect assembling of different links and lead to shifting and/or rotation of the frames associated with different elements, which are normally assumed to be matched and aligned. It is clear that the errors in geometrical parameters do not depend on the manipulator configuration, while their effect on the position accuracy depends on the last one. At present, there exists various sophisticated calibration techniques that are able to identify the differences between the actual and the nominal geometrical parameters [5-9]. Consequently, this type of errors can be efficiently

---

**Corresponding author:** Alexandr Klimchik, Ph.D. E-mail: alexandr.klimchik@mines-nantes.fr.

compensated either by adjusting the controller input (i.e. the target point coordinates) or by straightforward modification of the geometrical model parameters used in the robot controller.

In some other cases, the geometrical errors may be dominated by non-geometrical ones that may be caused by influences of a number of factors [10-11]. However, in the regular service conditions, the compliance errors are the most significant source of inaccuracy. Their influence is particularly important for heavy robots and for manipulators with low stiffness. For example, the cutting forces/torques from the technological process may induce significant deformations, which are not negligible in the precise machining. In this case, the influence of the compliance errors on the robot position accuracy can be even higher than the geometrical ones.

Generally, the compliance errors depend on two main factors: (i) the stiffness of the manipulator and (ii) the loading applied to it. Similar to the geometrical ones, the compliance errors highly depend on the manipulator configuration and essentially differ throughout the workspace [12]. So, in order to obtain correct prediction of the robot end-effector position, the efficient compliance errors compensation should be applied [13]. One way to solve this problem is to improve the accuracy of the stiffness model by means of elastostatic calibration. This procedure allows to identify the stiffness parameters from the redundant information on the robot end-effector position provided by the measurements, where the impacts of associated measurement noise on the calibration results have to be minimized by proper selection of measurement configurations.

However, currently most of the efforts have been made for kinematic calibration, only few works directly address the issue of elastostatic calibration and its influence on the robot accuracy [14]. In this area, using various manipulator configurations for different measurements seems to be also attractive and perfectly corresponds to some basic ideas of the classical design of experiments theory [15] that intends using the factors that are differed from each other as much as possible. In spite of potential advantages of this approach and potential benefits to improve the identification accuracy significantly, only few works addressed to the issue of the best measurement pose selection [16-19]. Hence, the problem of selection of the optimal measurement poses for elastostatic parameters calibration requires additional efforts. This problem can be treated as finding the strategy of determining a set of optimal measurement poses within the reachable joint space that minimize the effects of measurement noise on the estimation of the robot parameters. It should be mentioned that the end-effector location as well as its deflection under the loading are described by a non-linear set of functions. However, the classical results of the identification theory are mostly obtained for very specific models (such as linear regression). Therefore, they can not be applied directly and an additional enhancement is required.

One of the key issues in the experiment design theory is the comparison of different plans of experiment (i.e. sets of configurations and corresponding loadings). In the literature, in order to define the optimal plans of experiments, numerous quantitative performance measures have been proposed. They allow to define an optimization problem (either multiobjective or single-objective), whose solution yields the desired set of measurement poses [20-24]. However, all the existing performance measures have their limitations that affect the calibration accuracy in different manners. As a result, they do not entirely correspond to the industrial requirements.

In this paper, the problem of optimal design of the elastostatic calibration experiments is studied for the case of serial anthropomorphic manipulator, which obviously does not cover all architectures used in practice. Nevertheless, it allows us to derive very useful analytical expressions and to obtain some simple practical rules defining optimal configurations with respect to the calibration accuracy. In contrast to other works, a new criterion is proposed that evaluates the quality of compliance errors compensation based on the concept of manipulator test-pose. The proposed criterion has a clear physical meaning and is directly related to the robot accuracy under the task load. So, it aims at improving the efficiency of compliance errors compensation via proper selection of measurement poses.

The remainder of this paper is organized as follows. Section 2 addresses to the problem of elastostatic calibration, formulates basic assumptions and defines the research problem. Section 3 focuses on the influence of measurement errors on the identification accuracy. Section 4 proposes new test-pose based approach. In Section 5 the proposed approach is illustrated on the example of experiment design for the industrial robot. Finally, Section 6 summarizes the main results and contributions.

## 2.  Problem of elastostatic calibration

The elastostatic properties of a serial robotic manipulator [12] are usually defined by the Cartesian stiffness matrix $\mathbf{K}_C$, which is computed as

$$\mathbf{K_C} = \mathbf{J}^{-T}\mathbf{K_\theta}\mathbf{J}^{-1} \qquad (1)$$

where $\mathbf{J}$ is the Jacobian matrix with respect to the joint angles $\mathbf{q}$, and $\mathbf{K_\theta}$ is a diagonal matrix that aggregates the joint stiffness values. In order to describe the linear relation between the end-effector displacement and the external force, the stiffness model of this manipulator can be rewritten as follows

$$\Delta \mathbf{t} = \mathbf{J}\,\mathbf{k_\theta}\mathbf{J}^T\mathbf{W} \qquad (2)$$

where $\Delta \mathbf{t} = (\Delta \mathbf{p}^T, \Delta \boldsymbol{\varphi}^T)^T$ is the robot end-effector displacement (position $\Delta \mathbf{p}$ and orientation $\Delta \boldsymbol{\varphi}$) caused by the external loading $\mathbf{W}$, which includes the force $\mathbf{F}$ and torque $\mathbf{T}$ applied to the robot end-effector; $\mathbf{k_\theta}$ is the joints compliance matrix that is treated as an unknown below and should be identified from the calibration experiments.

In the scope of this paper, the following assumptions concerning the manipulator model and the measurement equipment limitations are accepted:

**A1**: It is assumed that the *geometric parameters are well calibrated*. So, for the unloaded mode ($\mathbf{W}=0$), the vector $\mathbf{q}$ is equal to the nominal value of the joint angles $\mathbf{q_0}$. However, for the case when the loading is not equal to zero $\mathbf{W} \neq 0$, the joint angles include deflections, i.e. $\mathbf{q} = \mathbf{q_0} + \Delta \mathbf{q}$, where $\Delta \mathbf{q}$ is the vector of joint displacements due to the external loading $\mathbf{W}$.

**A2**: It is assumed that *each calibration experiment produces three vectors* $\{\Delta \mathbf{p}_i, \mathbf{q}_i, \mathbf{W}_i\}$, which define the displacements of the robot end-effector, the corresponding joint angles and the external wrenches respectively, where $i$ is the experiment number. So, the calibration procedure may be treated as the best fitting of the experimental data $\{\Delta \mathbf{p}_i, \mathbf{q}_i, \mathbf{W}_i\}$ by using the stiffness model (2) that can be solved using the standard least-square technique.

**A3**: In practice, the calibration includes *measurements of the end-effector Cartesian coordinates with some errors*, which are assumed to be i.i.d (independent identically distributed) random values with zero expectation and standard deviation $\sigma$. Because of these errors, the desired values of $\mathbf{k_\theta}$ are always identified approximately.

Using these assumptions and the above defined notation, the problem of interest can be defined as:

**Problem:** To propose a technique for selecting the set of joint variables $\mathbf{q}_i$ and corresponding external wrench $\mathbf{W}_i$ for the elastostatic calibration of industrial robot that leads to the accuracy improvement for the given technological process.

Usually, the performance measures that evaluate the quality of the calibration plans are based on the analysis of the covariance matrix of the identified parameters, all elements of which should be as small as possible. However, in robotics, the stiffness parameters ($k_1, k_2, ...$) have different influences on the end-effector displacements; moreover, their influence varies throughout the workspace. To overcome this difficulty, it is assumed that:

**A4**: the *"calibration quality" is evaluated for the so-called test configuration* $\{\mathbf{q}_0, \mathbf{W}_0\}$, which is given by a user and for which it is required to have the best positioning accuracy under the external loading.

To obtain the optimal calibration plan of experiment for a typical industrial manipulator, two sub-problems should be considered: (i) to propose a performance measure for comparing different plans of experiments that are adopted to the elastostatic parameters calibration and are related to the robot accuracy under the task loading; (ii) to find optimal configurations of the manipulator for the elastostatic parameters calibration that provide the best compliance error compensation.

## 3. Influence of measurement errors

For computational convenience, the linear relation (2) where the desired parameters are arranged in the diagonal matrix $\mathbf{k_\theta} = diag(k_1, k_2, ...)$ should be rewritten in the following form

$$\Delta \mathbf{t}_i = \mathbf{A}_i \mathbf{k} \qquad (3)$$

where the vector $\mathbf{k}$ collects the joint compliances that are extracted from matrix $\mathbf{k_\theta}$. Here, the matrix $\mathbf{A}_i$ is defined by the columns of Jacobian $\mathbf{J}$ and the external force $\mathbf{F}$ and is expressed as

$$\mathbf{A}_i = \left[ \mathbf{J}_{1i}\mathbf{J}_{1i}^T\mathbf{W}_i, ..., \mathbf{J}_{ni}\mathbf{J}_{ni}^T\mathbf{W}_i \right] \quad (i = \overline{1,m}) \qquad (4)$$

where $\mathbf{J}_{ni}$ is the *n*-th column vector of the Jacobian matrix for the *i*-th experiment, *m* is the number of experiments. Using the identification theory, the joint compliances can be obtained from Eq. (3) using the least square method, which minimizes the residuals for all experimental data. The corresponding optimization problem

$$\sum_{i=1}^{m} (\mathbf{A}_i \mathbf{k} - \Delta \mathbf{t}_i)^T (\mathbf{A}_i \mathbf{k} - \Delta \mathbf{t}_i) \rightarrow \min_{\mathbf{q}_i, \mathbf{F}_i} \qquad (5)$$

provides the estimate of the desired parameters, which can be presented as

$$\hat{\mathbf{k}} = \left( \sum_{i=1}^{m} \mathbf{A}_i^T \mathbf{A}_i \right)^{-1} \cdot \left( \sum_{i=1}^{m} \mathbf{A}_i^T \Delta \mathbf{t}_i \right) \qquad (6)$$

However in practice, only translational deflections are measured directly. So, in order to reduce computational efforts, it is reasonable to eliminate equations that correspond to the rotational deflections from expression (3) and to rewrite it as

$$\Delta \mathbf{p}_i = \mathbf{A}_i^{(p)} \mathbf{k} \qquad (7)$$

where the matrix $\mathbf{A}_i^{(p)}$ corresponds to the position deflections only. For comparison, the original matrix from expression (3) includes an additional block $\mathbf{A}_i^{(\varphi)}$ corresponding to the rotational deflections:

$$\mathbf{A}_{i \ 6 \times n} = \begin{bmatrix} \mathbf{A}_{i \ 3 \times n}^{(p)} \\ \mathbf{A}_{i \ 3 \times n}^{(\varphi)} \end{bmatrix} \qquad (8)$$

So, expression (6) should be rewritten in the following form

$$\hat{\mathbf{k}} = \left( \sum_{i=1}^{m} \mathbf{A}_i^{(p)^T} \mathbf{A}_i^{(p)} \right)^{-1} \cdot \left( \sum_{i=1}^{m} \mathbf{A}_i^{(p)^T} \Delta \mathbf{p}_i \right) \qquad (9)$$

It is obvious that errors cannot be avoided in the calibration experiments. These errors mainly caused by the accuracy of the positioning measurement system while measuring the end-effector position can be expressed as

$$\Delta \mathbf{p}_i = \mathbf{A}_i^{(p)} \mathbf{k}_0 + \boldsymbol{\varepsilon}_i \qquad (10)$$

where $\mathbf{k}_0$ is the true value of the unknown parameter and $\boldsymbol{\varepsilon}_i$ is the measurement errors are assumed in the $i$-th experiment. Usually the errors are assumed to be independent identically distributed (i.i.d.) with zero expectation $E(\boldsymbol{\varepsilon}_i) = \mathbf{0}$ and the variance $E(\boldsymbol{\varepsilon}_i^T \boldsymbol{\varepsilon}_i) = \sigma^2$.

Using expression (10) the estimate of the compliance vector $\hat{\mathbf{k}}$ can be presented as

$$\hat{\mathbf{k}} = \mathbf{k}_0 + \left( \sum_{i=1}^{m} \mathbf{A}_i^{(p)^T} \mathbf{A}_i^{(p)} \right)^{-1} \left( \sum_{i=1}^{m} \mathbf{A}_i^{(p)^T} \boldsymbol{\varepsilon}_i \right) \qquad (11)$$

where the first term corresponds to the expectation $E(\hat{\mathbf{k}})$ (it means that the estimate (9) is unbiased).

It can be also proved that the covariance matrix of compliance parameters $\hat{\mathbf{k}}$ that defines the identification accuracy can be expressed as

$$\begin{aligned} \text{cov}(\hat{\mathbf{k}}) = \left( \sum_{i=1}^{m} \mathbf{A}_i^{(p)^T} \mathbf{A}_i^{(p)} \right)^{-1} & E \left( \sum_{i=1}^{m} \mathbf{A}_i^{(p)^T} \boldsymbol{\varepsilon}_i^T \boldsymbol{\varepsilon}_i \mathbf{A}_i^{(p)} \right) \\ & \times \left( \sum_{i=1}^{m} \mathbf{A}_i^{(p)^T} \mathbf{A}_i^{(p)} \right)^{-1} \end{aligned} \qquad (12)$$

and, taking into account that $E(\boldsymbol{\varepsilon}_i^T \boldsymbol{\varepsilon}_i) = \sigma^2$, it can be simplified to

$$\text{cov}(\hat{\mathbf{k}}) = \sigma^2 \left( \sum_{i=1}^{m} \mathbf{A}_i^{(p)^T} \mathbf{A}_i^{(p)} \right)^{-1} \qquad (13)$$

where $\sigma$ is the s.t.d. of the measurement errors. So, for the considered problem, the impact of the measurement errors is defined by the matrix sum $\sum_{i=1}^{m} \mathbf{A}_i^{(p)^T} \mathbf{A}_i^{(p)}$ that is also called the information matrix.

Obviously, in order to have the smallest dispersion of the identification errors, it is required to have the covariance matrix elements as small as possible. It is a multiobjective optimization problem, but the minimization of one element may increase others. In the literature, in order to reduce this problem to a monobjective one, numerous scalar criteria have been proposed. It should be mentioned that all these criteria provide rather different optimal solutions. Hence, it is quite important to select a proper optimization criterion that ensures the best position accuracy of the manipulator under the loading. For this reason, in the next section a new test-pose based approach that ensures the best end-effector accuracy under external loading is proposed.

## 4.    Test-pose-based approach

The main idea of the calibration experiment planning is to select proper configurations and corresponding external loadings (which will be called as plan of experiments) that ensure the best identification accuracy for the desired parameters. To develop this idea, let us introduce several definitions that are referred below to as D1, D2 and D3.

**D1**: *Plan of experiments* is a set of robot configurations and corresponding external loadings that are used for the measurements of the end-effector displacements and further identification of the elastostatic parameters.

As follows from previous works (mainly devoted to the geometrical calibration), proper selection of the plan of experiments allows us to achieve an essential reduction of the measurement error impact. However, there is an open question here that is related to the numerical evaluation of this impact. Corresponding expression can be treated as the objective function in the optimization problem, which produces the desired plan of experiments. It should be mentioned that for linear models this problem has been already carefully studied. In particular, in classical regression analysis, there are several conventional optimality criteria that operate with the trace and/or determinant of the covariance matrix or its inverse (so called information matrix). The most commonly used among them are presented in Table 1 and in conventional design of experiments [20-24] they are known as A-, D-, E-, G-optimality criteria.

In addition, in robot geometrical calibration that operates with non-linear models, some specific performance measures are used, which are based on the singular value decomposition of the kinematic Jacobian. This approach can be also adopted for the elastostatic calibration, where the SVD should be applied to the matrix $\mathbf{A}^{(p)}$, which contains both the kinematic Jacobian and the external loading vector. More details concerning these performance measures are presented in the second part of Table 1.

**Table 1  Objective function for existing approaches in calibration experiment design**

| Approach | Objective function |
|---|---|
| Application: Linear Regression | |
| A-optimality | $\mathrm{trace}(\mathrm{cov}(\mathbf{k})) \to \min\limits_{\mathbf{q}_i, \mathbf{W}_i}$ |
| D-optimality | $\det(\mathrm{cov}(\mathbf{k})^{-1}) \to \max\limits_{\mathbf{q}_i, \mathbf{W}_i}$ |
| E-optimality | $\min\{\mathrm{eig}(\mathrm{cov}(\mathbf{k})^{-1})\} \to \max\limits_{\mathbf{q}_i, \mathbf{W}_i}$ |
| G-optimality | $\max\{\mathrm{diag}(\mathbf{k})\} \to \min\limits_{\mathbf{q}_i, \mathbf{W}_i}$ |
| Application: Robot Calibration | |
| Product of singular values $O_1$ | $\sqrt[s]{\sigma_s \ldots \sigma_1} \to \max\limits_{\mathbf{q}_i, \mathbf{W}_i}$ |
| Condition number $O_2$ | $\sigma_1 / \sigma_s \to \min\limits_{\mathbf{q}_i, \mathbf{W}_i}$ |
| Minimum singular value $O_3$ | $\sigma_s \to \min\limits_{\mathbf{q}_i, \mathbf{W}_i}$ |
| Noise amplification index $O_4$ | $\sigma_s^2 / \sigma_1 \to \max\limits_{\mathbf{q}_i, \mathbf{W}_i}$ |
| Inverse sum of singular values $O_5$ | $\sum_i 1/\sigma_i \to \min\limits_{\mathbf{q}_i, \mathbf{W}_i}$ |
| $\sigma_1, \sigma_s$ maximum and minimum singular values | |

It should be mentioned that all optimization criteria, which are presented in Table 1, do not evaluate directly the measurement error impact on the robot accuracy in the technological application studied here. For this reason, in order to address the industrial requirements directly, it is proposed to estimate the quality of calibration experiment via the accuracy of the compliance error compensation. From statistical point of view, this approach can be treated as minimization of the prediction error. More strictly, an adopted performance measure is defined as follows:

**D2**: *The accuracy of the compliance error compensation $\rho$ is the distance between the desired end-effector location $\mathbf{t}_0$ and its real location under external loading $\mathbf{t}_F$ achieved after application of the compliance error compensation technique.*

Here, it is assumed that the desired end-effector location $\mathbf{t}_0$ is given or can be computed for given configuration $\mathbf{q}_0$ using manipulator direct geometrical model $g(...)$ as $\mathbf{t}_0 = g(\mathbf{q}_0)$. Since the external loading $\mathbf{W}$ causes the end-effector deflection with respect to the desired location, the compliance error compensation algorithm provides the modified values of the actuated coordinates $\mathbf{q} = \mathbf{q}_0 + \Delta\mathbf{q}$ that allow us to locate the end-effector at the given location $\mathbf{t}_0$. The letter can be expressed by the following equation

$$\mathbf{t}_0 = g(\mathbf{q}_0 + \Delta\mathbf{q}) + \mathbf{K}_C^{-1}\mathbf{W} \qquad (14)$$

where $\mathbf{K}_C$ is the Cartesian stiffness matrix computed for the configuration $\mathbf{q}_0$. Using linear approximation (assuming that the deflections are small enough), assuming that the Jacobian matrix is not singular the compliance error compensation algorithm can be presented as

$$\mathbf{q} = \mathbf{q}_0 - \mathbf{J}_0^{-1}\mathbf{K}_C^{-1}\mathbf{W} \qquad (15)$$

where $\mathbf{J}_0$ is the kinematic Jacobian computed for the same configuration $\mathbf{q}_0$. Geometrical interpretation of this algorithm is presented in Fig. 1, where three manipulator configurations are presented (the desired one as well as the configurations under the loading with and without compensation). In the case when the deflections are significant, the non-liner compliance error compensation technique should be applied [12].

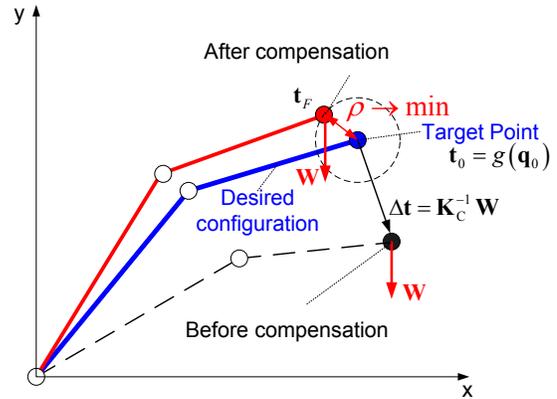

**Fig. 1   Geometrical interpretation of the compliance error compensation technique**

It should be noted that the compliance error compensation algorithm (15) includes the compliance matrix $\mathbf{K}_C^{-1}$, which is the function of the stochastic variables $\boldsymbol{\varepsilon}_i$ describing the measurement errors. For this reason, the desired compensation can be achieved "on average" only, while each particular case may produce some difference between the desired and compensated end-point locations (see Fig. 1).

Using notations from the previous section, the distance between the target and achieved locations may be computed as the Euclidean norm of $\delta\mathbf{p} = \mathbf{A}^{(p)} \cdot \delta\mathbf{k}$, where $\delta\mathbf{k} = \hat{\mathbf{k}} - \mathbf{k}_0$ is the difference between the estimated and true values of the robot stiffness parameters. It can be easily proved that the above presented algorithm (15) provides an unbiased compensation, i.e.

$$\mathrm{E}(\delta\mathbf{p}) = \mathbf{0} \qquad (16)$$

and the standard deviation of the compensation error $\rho^2 = \mathrm{E}(\delta\mathbf{p}^T\delta\mathbf{p})$ can be expressed as

$$\rho^2 = \mathrm{E}\left(\delta\mathbf{k}^T\mathbf{A}^{(p)T}\mathbf{A}^{(p)}\delta\mathbf{k}\right) \qquad (17)$$

Taking into account geometrical meaning of $\rho$, this value can be used as a numerical measure of the compliance error compensation quality (and also as a quality measure of the related plan of calibration experiments).

It is obvious that because of non-homogeneity of the manipulator properties within the workspace, the accuracy of the compliance error compensation highly depends on the target point location $\mathbf{t}_0$ and the applied external loading $\mathbf{W}$. For this reason, it cannot be evaluated in general for the whole robot workspace and variety of external loadings. To overcome this difficulty, it is proposed here to assess the compliance error compensation accuracy for some given manipulator configuration and typical external loading. This idea is formalized in the notion of the "test pose" defined below:

**D3**: The *test-pose* is the set of the robot configuration $\mathbf{q}_0$ and corresponding external loading $\mathbf{W}_0$ for which it is required to achieve the best compliance error compensation (i.e. $\rho_0^2 \to \min$).

Below, the test pose will be defined via the matrix $\mathbf{A}_0^{(p)}$, which is computed using expression (4). In practice, the values of $\mathbf{q}_0$ and $\mathbf{W}_0$ are provided by the user and usually correspond to a typical robot posture and cutting force for considered technological application. From this point of view, $\rho_0$ is treated as a measure of the robot accuracy in the machining process.

In the frame of the adopted notations, the proposed performance measure $\rho_0^2$ that evaluates the efficiency to compensate the compliance errors for the given test pose can be expressed as

$$\rho_0^2 = \mathrm{E}\left(\delta\mathbf{k}^T\mathbf{A}_0^{(p)T}\mathbf{A}_0^{(p)}\delta\mathbf{k}\right), \qquad (18)$$

where $\delta\mathbf{k} = \hat{\mathbf{k}} - \mathbf{k}_0$ is the elastostatic parameters estimation error caused by the measurement noise. This expression can be simplified by presenting the term $\delta\mathbf{p}^T\delta\mathbf{p}$ as the trace of the matrix $\delta\mathbf{p}\delta\mathbf{p}^T$, which yields

$$\rho_0^2 = \mathrm{trace}\left(\mathbf{A}_0^{(p)}\,\mathrm{E}\left(\delta\mathbf{k}\delta\mathbf{k}^T\right)\mathbf{A}_0^{(p)T}\right) \qquad (19)$$

Further, taking into account that $\mathrm{E}(\delta\mathbf{k}\delta\mathbf{k}^T)$ is the covariance matrix of desired parameters estimates $\hat{\mathbf{k}}$, the proposed performance measure (18) can be presented in the final form as

$$\rho_0^2 = \sigma^2\,\mathrm{trace}\left(\mathbf{A}_0^{(p)}\left(\sum_{i=1}^m \mathbf{A}_i^{(p)T}\mathbf{A}_i^{(p)}\right)^{-1}\mathbf{A}_0^{(p)T}\right) \qquad (20)$$

As follows from this expression, the proposed performance measure $\rho_0^2$ can be treated as the weighted trace of the covariance matrix $\mathrm{cov}(\hat{\mathbf{k}})$, where the weighting coefficients are obtained using the test pose. It has obvious advantages compared to previous approaches, which operate with "pure" trace of the covariance matrix (see Table 1) and involve straightforward summing of the covariance matrix diagonal elements, which may be of different units (corresponding to rotational and translational compliances, for instance). It should be noted that for the geometrical calibration, a similar approach has been used in [25].

Based on this performance measure, the calibration experiment design can be reduced to the following optimization problem

$$\mathrm{trace}\left(\mathbf{A}_0^{(p)}\left(\sum_{i=1}^m \mathbf{A}_i^T\mathbf{A}_i\right)^{-1}\mathbf{A}_0^{(p)T}\right) \to \min_{\{\mathbf{q}_i,\mathbf{F}_i\}} \qquad (21)$$

subject to

$$\|\mathbf{F}_i\| < F_{\max}, \qquad i = 1..m \qquad (22)$$

whose solution gives a set of the desired manipulator configurations and corresponding external loadings. It is evident that its analytical solution can hardly be obtained and a numerical approach is the only reasonable one.

Hence, the proposed above test-pose-based approach and related optimization problem ensure low values of the covariance matrix elements and allows to combine multiple objectives with different units in a single scalar objective. An application of this approach for the design of the calibration experiments is illustrated in the next section.

## 5.   Calibration experiment design for 6 d.o.f. manipulator: KUKA KR-270

Now let us consider the example that deals with calibration experiments design for the industrial robot KUKA KR-270 (Fig. 2). This robot has six actuated joints, which are assumed to be flexible. The links of the robot are quite stiff and are considered as rigid.

For such a manipulator (where the first joint defines the robot orientation in the xy-plane), it is reasonable to decompose the elastostatic calibration into two independent steps. The first step includes calibration of stiffness coefficients for joints 2...6 with vertical loading only. The second step includes the stiffness parameter calibration for the joint 1. It is obvious that the second step is quite easy from the experiment design point of view. In this case the optimization problem has only one variable for each configuration and the classical experiment design theory can be applied directly. In contrast, the first step is non trivial and requires intensive computations (corresponding results are presented below).

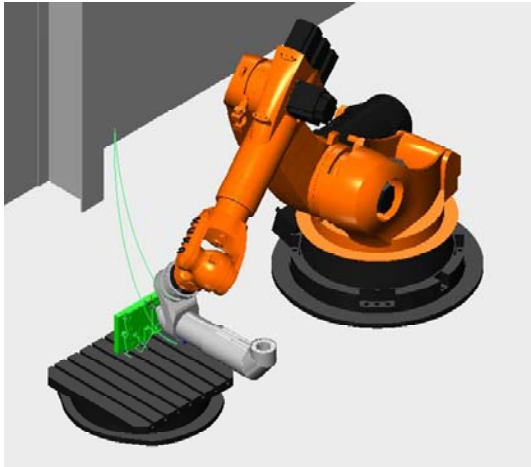

**Fig. 2     Machining configuration for the robot Kuka KR-270 (Test pose)**

In more details, the geometrical model and parameters of the robot are presented in Fig. 3 and Table 2, which also contains definition of the test pose that is presented in Fig. 2) [26].

It should be noted that for the machining process and for the elastostatic calibration different tools are used (see CAD models presented in Fig. 4). For this reason, computation of the matrices $\mathbf{A}_0^{(p)}$ and $\mathbf{A}_t^{(p)}$ involves different geometrical transformations "Tool". For given test configuration, the first of these matrices is defined as follows

$$\mathbf{A}_0^{(p)} = \begin{bmatrix} -73.4 & -177.4 & -106.1 & 102.4 & 0 \\ 0 & 0 & 197.2 & 19.3 & 0 \\ -363.6 & -98.3 & -167.1 & -42.2 & 0 \end{bmatrix} \quad (23)$$

**Table 2     Initial data for robotic-based milling**

| Test configuration, [deg] | | | | | |
|---|---|---|---|---|---|
| $q_1$ | $q_2$ | $q_3$ | $q_4$ | $q_5$ | $q_6$ |
| 75 | -56.9 | 89.3 | 45.1 | 76 | 57.2 |
| Machining force, [N] and torque [N m] | | | | | |
| $F_x$ | $F_y$ | $F_z$ | $T_x$ | $T_y$ | $T_z$ |
| 0 | 280 | -180 | 0 | 0 | 0 |
| Geometrical parameters, [mm] | | | | | |
| $d_1$ | $d_2$ | $d_3$ | $d_4$ | $d_5$ | $d_6$ |
| 350 | 750 | 1250 | -55 | 1100 | 0 |

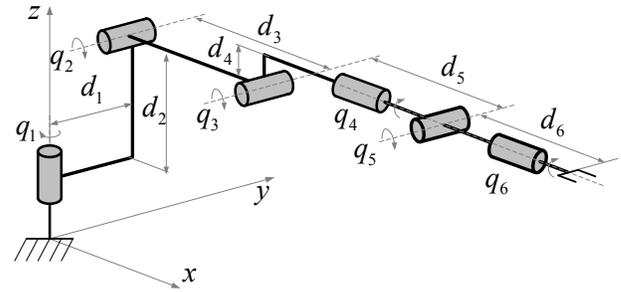

**Fig. 3     Geometrical model of Kuka KR-270**

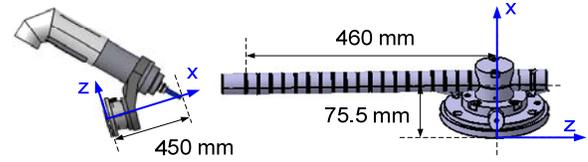

(a) machining tool          (b) tool for experiments

**Fig. 4     Tools used for machining and elastostatic calibration**

For the considered application example, there is a number of very specific constraints that are usually not considered in pure theoretical studies. In particular, there is a number of obstacles in the robot workspace (Fig. 5) that do not allow to achieve some configurations and to apply forces in some directions (vertical payload is obviously preferable). These constraints are summarized in Table 3. In addition, it is necessary to take into account usual constraints of the range of the joint variables ("joint limits").

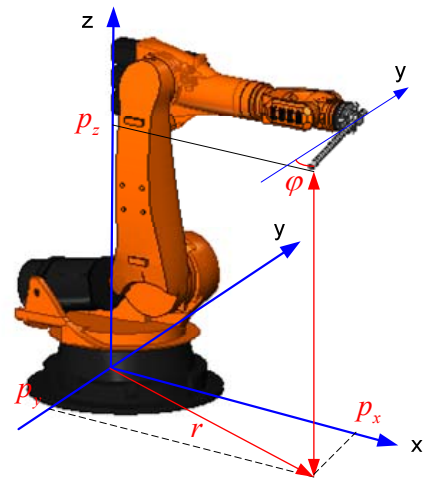

**Fig. 5     Workspace parameters included in the constraints for the elastostatic calibration**

**Table 3     Geometrical constraints for the elastostatic calibration**

| $p_z$ | $r$ | $|\varphi|$ |
|---|---|---|
| $> 800\,mm$ | $> 600\,mm$ | $< \pi/6$ |

**Table 4        Measurement configurations for the elastostatic calibration**

| $\mathbf{q}_i$ | Measurement configurations, [deg] | | | | |
|---|---|---|---|---|---|
| | $q_2$ | $q_3$ | $q_4$ | $q_5$ | $q_6$ |
| 2 calibration experiments | | | | | |
| $\mathbf{q}_1$ | -99.9 | 114.3 | -48.5 | 28.1 | -180 |
| $\mathbf{q}_2$ | -67.8 | -94.0 | 137.5 | -111.9 | 76.9 |
| 3 calibration experiments | | | | | |
| $\mathbf{q}_1$ | -93.5 | 125.0 | -118.7 | -62.4 | -168.9 |
| $\mathbf{q}_2$ | -103.4 | 93.4 | -147.9 | 105.8 | 93.2 |
| $\mathbf{q}_3$ | -98.9 | 113.8 | 50.6 | -38.6 | 16.8 |
| 4 calibration experiments | | | | | |
| $\mathbf{q}_1$ | -81.1 | 64.9 | -55.4 | 42.2 | 149.7 |
| $\mathbf{q}_2$ | -96.6 | 15.4 | 112.1 | -19.7 | 178.4 |
| $\mathbf{q}_3$ | -111.2 | -69.0 | 133.5 | 113.9 | -118.8 |
| $\mathbf{q}_4$ | -108.1 | 93.5 | -34.2 | -108.9 | 73.3 |
| 6 calibration experiments | | | | | |
| $\mathbf{q}_1$ | -84.0 | 126.9 | -119.4 | -61.2 | -172.9 |
| $\mathbf{q}_2$ | -105.5 | 98.6 | -148.2 | 99.7 | 94.7 |
| $\mathbf{q}_3$ | -106.0 | 106.8 | 49.4 | -38.5 | 22.2 |
| $\mathbf{q}_4$ | -89.2 | 132.3 | -119.2 | -61.4 | -173.9 |
| $\mathbf{q}_5$ | -96.7 | 86.6 | -147.1 | 102.3 | 96.0 |
| $\mathbf{q}_6$ | -99.4 | 108.8 | 51.8 | -39.1 | 17.6 |
| 12 calibration experiments | | | | | |
| $\mathbf{q}_1$ | -83.8 | 127.6 | -120.0 | -60.9 | -173.8 |
| $\mathbf{q}_2$ | -105.9 | 99.1 | -148.5 | 100.1 | 94.7 |
| $\mathbf{q}_3$ | -105.7 | 107.1 | 49.4 | -39.1 | 22.1 |
| $\mathbf{q}_4$ | -89.4 | 131.6 | -119.0 | -61.4 | -172.8 |
| $\mathbf{q}_5$ | -97.1 | 85.8 | -146.8 | 101.5 | 96.1 |
| $\mathbf{q}_6$ | -99.4 | 107.8 | 52.8 | -39.9 | 17.6 |
| $\mathbf{q}_7$ | -83.3 | 126.1 | -118.9 | -60.3 | -171.9 |
| $\mathbf{q}_8$ | -106.2 | 98.1 | -148.0 | 99.7 | 95.1 |
| $\mathbf{q}_9$ | -106.1 | 106.5 | 49.6 | -38.1 | 21.6 |
| $\mathbf{q}_{10}$ | -89.8 | 133.4 | -119.0 | -60.8 | -174.0 |
| $\mathbf{q}_{11}$ | -97.6 | 85.9 | -146.3 | 102.7 | 96.3 |
| $\mathbf{q}_{12}$ | -98.9 | 109.6 | 52.6 | -39.5 | 18.2 |

**Table 5        Elastostatic parameters estimation error**

| Number of exp. | Estimation error, [rad/ N m×10$^{-9}$] | | | | |
|---|---|---|---|---|---|
| | $\delta k_2$ | $\delta k_3$ | $\delta k_4$ | $\delta k_5$ | $\delta k_6$ |
| 2 exp. | 6.55 | 6.88 | 24.0 | 34.5 | 71.9 |
| 3 exp. | 5.74 | 6.87 | 19.2 | 26.4 | 74.9 |
| 4 exp. | 3.72 | 6.96 | 16.9 | 21.2 | 66.9 |
| 6 exp. | 3.93 | 4.82 | 13.8 | 16.4 | 55.2 |
| 12 exp. | 2.78 | 3.41 | 9.75 | 11.6 | 38.8 |

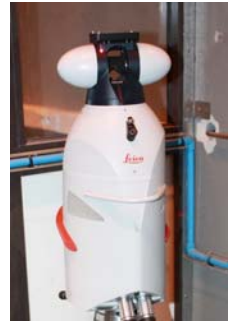

**Fig. 6        FARO laser tracker**

For this setting, it was solved the optimization problem (21) which produced the calibration experiment plans for $m \in \{2,3,4,6,12\}$. While solving this problem, it was assumed that the end-effector position was estimated using the FARO laser tracker (Fig. 6) [27], for which the measurement errors can be presented as unbiased random values with s.t.d. $\sigma = 0.03\,mm$. It is also assumed that the applied loading is the same for all calibration experiments and is equal to $\mathbf{F}_i = [0,\ 0,\ -2500, 0, 0, 0]^T$. The letter allows us to reduce the number of design variables by the factor of two. For the computations the workstation Dell Precision T7500 with two processors Intel® Xeon® X5690 (Six Core, 3.46GHz, 12MB Cache12) and 48 GB 1333MHz DDR3 ECC RDIMM was used. Since the optimisation problem (21) is quite sensitive to the starting point, parallel computing with huge number of the initial points were used.

The obtained results are summarized in Tables 4, 5 and 6. They include the identification errors for the elastostatic parameters, the accuracy of the error compensation $\rho_0$ for different plans of experiments and detailed descriptions of the measurement configurations. Table 6 also includes some additional results obtained by multiplication of the measurement configurations, which show that it is not reasonable to solve optimization problem for 12 configurations (that produce 60 design variables). However, almost the same accuracy of the compliance error compensation can be achieved by carrying out 12 measurements in 3 different configurations only (4 measurements in each configuration). This conclusion is in good agreement with the results presented in the previous section for 3 d.o.f. manipulator.

For comparison purposes, Fig. 7 presents simulation results obtained for different types of calibration experiments. As follows from them, any optimal plan (obtained for the case of two, three, four, six or twelve calibration experiments) improves the accuracy of the compliance error compensation in the given test pose by about 60% comparing to the random plan. Also, it is illustrated that repeating experiments with optimal plans obtained for the lower number of experiments provides almost the same accuracy as "full-dimensional" optimal plan. Obviously, the reduction of the measurement pose number is very attractive for the engineering practice.

## 6. Conclusions

The paper presents a new approach for the design of the elastostatic calibration experiments for robotic manipulators that allows essentially reducing the identification errors due to proper selection of the manipulator configurations and corresponding loadings, which are used for the measurements. In contrast to other works, the quality of the plan of experiments is estimated using a new performance measure that evaluates the efficiency of the compliance error compensation in the given test-pose. This approach allows to combine multiple objectives with different units in a single performance measure and ensures the best position accuracy for the given test configuration under the task loading. The proposed criterion can be treated as the weighted trace of the covariance matrix, where the weighting coefficients are derived using the test pose parameters.

The advantages of the developed technique are illustrated by an examples that deal with the calibration experiment design for 6 d.o.f. manipulator. It shows the benefits of the proposed approach, which is expressed via the position accuracy under the task loading.

**Table 6** The accuracy of the error compensation $\rho_0$ for different plans of experiments, [mm×$10^{-3}$]

| Number of exp. | Number of different configuration | | | | |
|---|---|---|---|---|---|
| | 2 conf. | 3 conf. | 4 conf. | 6 conf. | 12 conf. |
| 2 exp. | 5.989 | | | | |
| 3 exp. | --- | 4.676 | | | |
| 4 exp. | 4.235 (4.72%) | --- | 4.044 | | |
| 6 exp. | 3.458 (7.13%) | 3.306 (2.42%) | --- | 3.228 | |
| 12 exp. | 2.445 (7.14%) | 2.338 (2.45%) | 2.335 (2.32%) | 2.283 (<0.01%) | 2.282 |

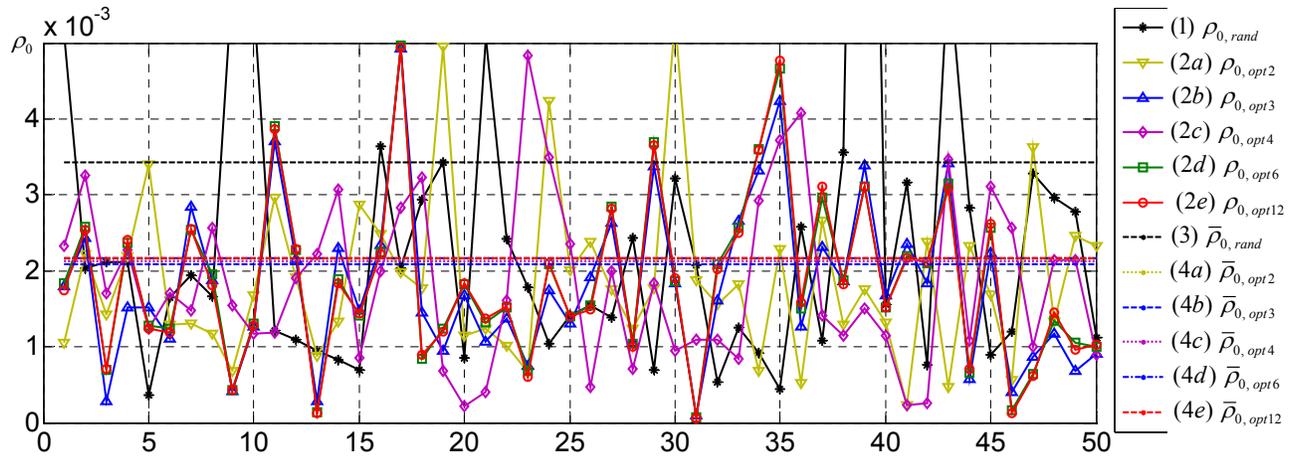

**Fig. 7** The accuracy of the compliance error compensation for different plans of calibration experiments for Kuka KR-270 manipulator for $\sigma = 0.03\,mm$ : **(1)** random plan $\rho_{0,\,rand}$ ; **(2a)** six experiments for optimal plan obtained for two calibration experiment $\rho_{0,\,opt2}$ , **(2b)** four experiments for optimal plan obtained for three calibration experiment $\rho_{0,\,opt3}$ , **(2c)** three experiments for optimal plan obtained for four calibration experiment, $\rho_{0,\,opt4}$ , **(2d)** two experiments for optimal plan obtained for six calibration experiment, $\rho_{0,\,opt6}$ , **(2e)** experiments for optimal plan obtained for twelve calibration experiment $\rho_{0,\,opt12}$ ; **(3)** expectation for plan **(1)** $\overline{\rho}_{0,\,rand} = 3.43{\cdot}10^{-3}\,mm$ ; **(4a)** expectation for plan **(2a)** $\overline{\rho}_{0,\,opt2} = 2.15{\cdot}10^{-3}\,mm$ ; **(4b)** expectation for plan **(2b)** $\overline{\rho}_{0,\,opt3} = 2.09{\cdot}10^{-3}\,mm$ ; **(4c)** expectation for plan **(2c)** $\overline{\rho}_{0,\,opt4} = 2.13{\cdot}10^{-3}\,mm$ ; **(4d)** expectation for plan **(2d)** $\overline{\rho}_{0,\,opt6} = 2.17{\cdot}10^{-3}\,mm$ ; **(4e)** expectation for plan **(2e)** $\overline{\rho}_{0,\,opt12} = 2.16{\cdot}10^{-3}\,mm$ ;

Besides, the results show that the combination of the low-dimension optimal plans gives almost the same accuracy as a full-dimensional plan. This conclusion allows the user to reduce essentially the computational complexity required for the calibration experiment design.

In future, the proposed approach will be extended for the case of simultaneous calibration of geometrical and elastostatic parameters. Another problem, which requires additional investigation is the experiment design for the set of the test poses (or for a long machining trajectory).

## Acknowledgements

The work presented in this paper was partially funded by the ANR, France (Project ANR-2010-SEGI-003-02-COROUSSO).